\documentclass[conference,10pt]{IEEEtran} \IEEEoverridecommandlockouts
\usepackage[noadjust]{cite}
\usepackage{amsmath,amssymb,amsfonts}
\usepackage{tabularx}
\usepackage{graphicx}
\usepackage{subcaption}
\usepackage{booktabs}
\usepackage{textcomp}
\usepackage{xcolor}
\usepackage[hidelinks]{hyperref}
\usepackage[linesnumbered]{algorithm2e}
\usepackage{algorithmic}
\DeclareMathOperator*{\argmin}{argmin}

\DeclareMathOperator*{\linspace}{linspace}

\def\BibTeX{{\rm B\kern-.05em{\sc i\kern-.025em b}\kern-.08em
    T\kern-.1667em\lower.7ex\hbox{E}\kern-.125emX}}
\newcommand{\keyfinding}[1]{\noindent\fbox{\parbox{0.97\linewidth}{\textbf{Key findings}: #1}}}

\begin{document}

\title{Understanding The Effectiveness of Lossy Compression in Machine Learning Training Sets\\
\thanks{This paper has been supported funding from the National Science Foundation and the US Department of Energy}
}

\author{Robert Underwood, Jon C. Calhoun
\IEEEmembership{Senior Member,IEEE}, Sheng Di \IEEEmembership{Senior Member,IEEE}, Franck Cappello \IEEEmembership{Fellow,IEEE}\thanks{R. Underwood, S. Di, and F. Cappello are with Argonne National Laboratory and J. Calhoun is with Clemson University}}

\markboth{IEEE Cluster, May 2023}{BLIND: Understanding The Effectiveness of Lossy Compression in Machine Learning Training Sets}

\maketitle

\begin{abstract}
Machine Learning and Artificial Intelligence (ML/AI) techniques have become increasingly prevalent in high performance computing (HPC).  However, these methods depend on vast volumes of floating point data for training and validation which need methods to share the data on a wide area network (WAN) or  to  transfer  it  from  edge devices to  data  centers.  Data compression can be a solution to these problems, but an in-depth understanding of how lossy compression affects model quality is needed.  Prior work largely considers a single application or compression method.  We designed a systematic methodology for evaluating data reduction techniques for ML/AI, and  we use it to perform a very comprehensive evaluation with 17 data reduction methods on 7 ML/AI applications to show modern lossy compression methods can achieve a 50-100$\times$ compression ratio improvement  for a 1\% or less loss in quality.  We identify critical insights that guide the future use and design of lossy compressors for ML/AI.
\end{abstract}


\section{Introduction}

\IEEEPARstart{I}{n} recent years, there has been an increased focus on the use of machine learning  techniques and artificial intelligence (ML/AI) in high performance computing (HPC) applications in many domains \cite{babuModelBasedSemanticCompression,isakovHPCThroughputBottleneck2020,wozniakCANDLESupervisorWorkflow2018} that require huge amounts of data for training.
The Candle project -- which does cancer research -- alone anticipates more than \textit{1PB} of floating point data required for training \textit{per experiment}, for just \textit{one class} of problem they consider \cite{wozniakCANDLESupervisorWorkflow2018}. Additionally, advanced instruments such as Linac Coherent Light Source (LCLS-II) \cite{LCLSII} and Advanced Photon Source (APS) \cite{APSU} may produce the floating-point X-ray imaging data at a data acquisition rate exceeding 1TB/s. 
Smaller floating point data streams also struggle in real-world uses with limited bandwidth.
Emerging HPC use cases such as  structural health monitoring \cite{lockeUsingDrivebyHealth2020} or traffic safety applications \cite{rahmanDynamicErrorboundedLossy} move floating point data from the edge to nearby HPC centers for processing in parallel.
These applications need data transferred over rural cellular networks to an HPC center for near real-time processing.
With these techniques comes a desire for processing an increasing volume of floating point data for training to improve the ability to store and transport the data efficiently.

Applications largely consider data reduction techniques of training data for three reasons:
\textit{1)  To accelerate WAN transfer times and reduce network and storage costs and time for reproducibility}.
Reproducibility is key to the scientific enterprise, and it often requires transporting datasets to other sites because of limitations on compute availability and other usage policy reasons.
Downloads of the Pile dataset (800GB) used to train LLMs generated nearly 320TB of traffic last month alone on HuggingFace likely costing many thousands of dollars in bandwidth costs alone.
Beyond the cost, in developing countries, it would take nearly 2 days assuming no contention or failures to download the Pile once.
Doing the same for a 5-minute experiment just once from the upgraded APS or LCLS-II-HE would take nearly 2 years and cost between 10 to 20 thousand USD on commercial cloud providers.
\textit{2) Reducing equipment costs by transporting data}
To allow consolidation of GPUs and other high-cost computing equipment centrally instead of deploying them in many IoT/Edge devices to reduce costs such as in intelligent transportation systems in rural networks as done in \cite{rahmanDynamicErrorboundedLossy}.  A similar use case applies to structural health monitoring for bridges in rural environments \cite{lockeUsingDrivebyHealth2020}
\textit{3) Improving performance by storing full datasets locally}
To allow storing full datasets on nodes that otherwise do not fit in memory or on local SSD storage.  The 1.6TB ROOT dataset exceeds the local SSD on Frontier when also storing checkpoints and the optimizer state of the model. 3 seconds of data from the upgraded APS \cite{APSU} also would exhaust local storage.

\begin{figure*}
    \centering
    \includegraphics[width=\textwidth]{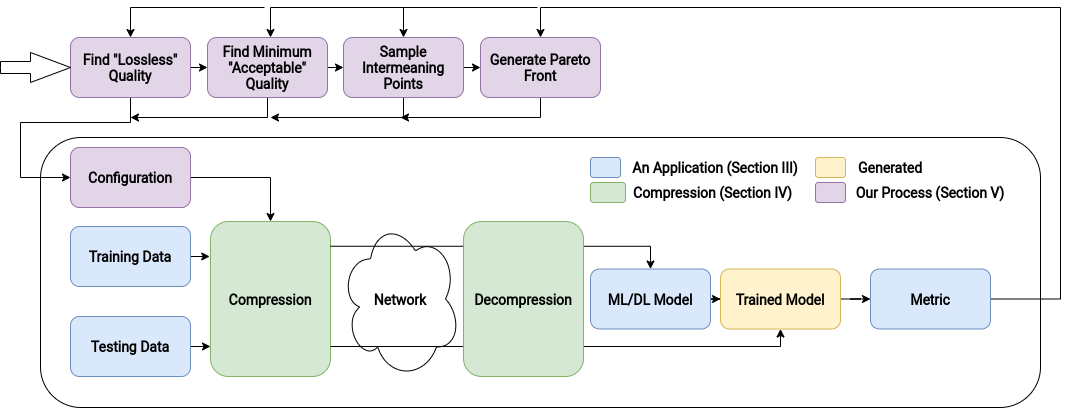}
    \caption{Workflow Overview}
    \label{fig:mlai:workflow}
\end{figure*}

The challenges of storing/transferring data for scientific applications have led researchers to consider using lossy compression methods to significantly reduce the data volumes. Compared with lossless compression which achieves only  modest compression ratios for scientific floating-point data-sets \cite{cappelloUseCasesLossy2019}, lossy compression techniques \cite{cappelloUseCasesLossy2019, diFastErrorBoundedLossy2016, lindstromFixedRateCompressedFloatingPoint2014} yield larger compression ratios. 

For lossy compression, the most critical concern for users is to understand the impact of the information loss on the quality of the results of the application.
Without a good understanding of such impact, it is very difficult to determine the most efficient/qualified lossy compressor with specific fidelity requirements on the quality metrics for applications.
There have been relatively few studies that look at the effect of lossy compression of training data on quality for ML/AI applications \cite{rahmanDynamicErrorboundedLossy}.
These studies were limited in that they consider relatively few methods often on a single application and do not propose a general methodology for comparing compression methods or applications.

Our contributions are as follows:

\begin{enumerate}
    \item A methodology for evaluating the uses of Lossy Compression of training data for ML/AI applications.
    \item We perform a more comprehensive evaluation of 17+ data reduction methods on training data of 7 ML/AI applications; more than has previously been attempted.
    \item We demonstrate that lossy compression is both safe and effective on a range of ML/AI applications achieving a 50-100$\times$ improvement in compression ratio for 1\% or less decrease in quality with greater improvements possible at lower quality on a variety of applications.
    \item We demonstrate that data compression is best performed with a value range relative error bound independently on each column on tabular floating point data.
    \item We describe a method to candidate Pareto points of configuration to more effectively sample the search space of configurations to identify possible solutions to the multi-objective optimization problem of compressor configuration
\end{enumerate}

Figure~\ref{fig:mlai:workflow} summarizes our proposed methodology that we expand upon throughout the remainder of our paper.
In Section~\ref{sec:mlai:applications} we define our concept of an application and provide background on the applications we consider and how their developers define quality (indicated by the blue in Figure~\ref{fig:mlai:workflow}).
In Section~\ref{sec:mlai:background}, we discuss the compression methods used in our evaluation (green).
We evaluate each application several times with different compressors and configurations of those compressors to understand the trade-offs.
The process of evaluating applications is very expensive so in Section~\ref{sec:mlai:methodology}, we describe an approach to identify configurations with useful trade-offs that impact on  user's application analysis results (purple).
After that, we present results that show the trade-off in quality using our method and describe the insights for use of lossy compression in ML/AI and the future development of compressors.
Last, we discuss how to use parallel compression to meet bandwidth constraints to achieve a speedup for training and offer final conclusions.

\section{Applications}\label{sec:mlai:applications}
For the purpose of this paper, a problem (represented in blue in Figure~\ref{fig:mlai:workflow}) is some specific use case for ML/AI such as predicting if a particular drug treats a particular kind of tumor (i.e. the P:Drug problem from Candle).
An application is a particular solution to the problem (i.e. Candle NT-3).
Therefore there are many applications for a particular problem.
An application consists of a training and a testing dataset, code that applies ML/AI to the application (a model), and at least one metric that determines how well the model performs on the problem (such as accuracy, Pearson $R^2$ coefficient, etc...).

To investigate the impacts of applying compression to ML/AI, we treat the applications \textit{as a control variable} for the purpose of the paper -- that is we reused \textit{existing} applications that perform well without compression according to the metric for their particular problem \textit{while just introducing compression}.
We choose these applications because they are publicly available examples of ML/AI performed on floating-point data in the context of HPC.
This excludes other problems that are better studied such as compression for non-scientific images or video.

We consider the application of compression to two possible points of the training and evaluation process.
We consider decompressing the training data prior to training and decompressing the validation data prior to validation.
This model is consistent with storing or receiving the data in compressed form prior to training or validating on it.
While we are focusing on training and validation data in this paper, we expect this methodology to extend to other points of applying compression.

Table~\ref{tab:data-sets} summarizes each of the problems that we consider in our work.
For each approach, we perform no additional normalization or prepossessing of features beyond what is provided by the applications we describe below. 

\begin{table*}[t]
    \begin{tabularx}{\textwidth}{lcccclXl}
        \toprule
            Dataset & Size & Observations & Features & Data Type & Problem Solved & Approach & Author's Metric \\
        \midrule
            Superconductor \cite{hamidiehDataDrivenStatistical2018} & 14Mb & 21263 & 82 & Real & Regression & Random \mbox{Forest} & $R^2$ \\
            MLSVM-Advertisement \cite{sadrfaridpourEngineeringFastMultilevel2019} & 59MB & 3279 & 1558 & Real & Binary \mbox{Classification} & MLSVM (SVM) & G-Mean \\
            MLSVM-Buzz \cite{sadrfaridpourEngineeringFastMultilevel2019} & 125MB & 140707 & 77 & Real & Binary \mbox{Classification} & MLSVM (SVM) & G-Mean \\
            Candle-NT3 \cite{wozniakCANDLESupervisorWorkflow2018} (Small) & 518Mb & 1120 & 60483 & Real & Binary \mbox{Classification} & Deep Convolutional Network  & Validation Accuracy\\
            Candle-NT3 \cite{wozniakCANDLESupervisorWorkflow2018} (Large) & 4Gb & 8960 & 60483 & Real & Binary \mbox{Classification} & Deep Convolutional Network & Validation Accuracy \\
            Ptychonn \cite{cherukaraAIenabledHighresolutionScanning2020} (Large)  & 4Gb & 8100 & 131072 & Real & \mbox{Data Reconstruction} & Convolutional \mbox{Encoder}/\mbox{Decoder} & MSE/PSNR \\
        \bottomrule
    \end{tabularx}
    \caption{Overview of the Applications}
    \label{tab:data-sets}
\end{table*}

Superconductor \cite{hamidiehDataDrivenStatistical2018} uses features from various materials to predict the critical temperature where the material becomes superconductive for various materials.  The application we identified used a random forest regressor.  The authors use the standard Pearson's $R^2$ metric to evaluate their results which measures the strength of the explanatory power of the model.  For $R^2$ the minimum value is 0, the maximum value is 1, and larger values are better. We include it as a classic, science problem, and as an example of a regression problem.

MLSVM \cite{sadrfaridpourEngineeringFastMultilevel2019} is an implementation of support vector machines.  Support Vector Machines are a widely used and effective tool in ML/AI.  In each case, the algorithm performs binary classification.  The authors feature in their evaluation several tuning parameters for their algorithm.  We use the tuning that the authors identified  were best on the uncompressed inputs for these parameters in their paper to make as direct a comparison as possible.  The authors evaluate their results with the geometric mean of precision and recall because many of these problems display a degree of imbalance. It has a minimum value of 0, a maximum value of 1, and larger values are better  We include it as some of the larger data sets we identified for classic machine learning problems.

Candle NT3 aims to solve the problem of detecting if a particular drug will treat a particular kind of tumor.  Candle addresses this problem with a 1d deep convolutional network.
The authors choose to evaluate their work with the validation accuracy \cite{wozniakCANDLESupervisorWorkflow2018} \footnote{ Other metrics could be used, but we follow the author's choice because we treat applications as control variables.  In a medical domain metrics like precision or sensitive are more common}.
This metric has a mimimum of 0, a maximum of 1, and larger values are better.
Candle represents a larger publicly available dataset than would identify with classic problems.  Additionally, Candle expects to have a massive 1PB for each run of this problem in the future.  The Candle project provides a data scaling algorithm to synthesize the larger data sets for run-time bench-marking purposes however there is not yet an evaluation of its effect on the quality of predictions \cite{wozniakCANDLESupervisorWorkflow2018}.

Ptychonn uses an encoder and decoder to reconstruct X-ray images from sensor data from the APS.
The authors evaluate their work using the Mean Squared Error (MSE) of the reconstructed image with a ground truth classical but computationally expensive approach.
The MSE has a minimum of 0, has an unbounded maximum, and smaller values are better.

Our preliminary work also considered 8 other data sets from the UCI Machine Learning Database \cite{duaUCIMachineLearning2017} that we omit here for space and because they have smaller data volume ($<15$MB), but observe similar results on these data sets.

\section{Data Reduction Techniques Studied} \label{sec:mlai:background}

We turn our attention to which compression techniques to evaluate.
Table~\ref{tab:compression_methods} summarizes the methods we considered in this paper.
In order to understand the impact of lossy compression to ML/AI analysis results, it is necessary to have an  understanding of the various data reduction techniques' design principles as well as their pros and cons. To this end, we classify and describe the 17 data reduction techniques we considered in this section. All the techniques mentioned here are studied and compared in our analysis in Section~\ref{sec:mlai:results}. 
It is important to include so many different methods because different sub-communities of ML/AI use different approaches for data reduction in order to understand relative the improvements offered by lossy compression, and because different lossy methods induce different kinds of errors which may effect applications differently.

\begin{table}[t]
    \centering
    \caption{Summary of Data Reduction Methods}
    \label{tab:compression_methods}
    \begin{tabularx}{\columnwidth}{lX}
        \toprule
        Method & Description \\
        \midrule
        GZIP & \cite{deutschRFC1952GZIP1996} lossless compressor known for its compression ratios; used via cBLOSC \\
        LZ4 & \cite{colletLZ4ExtremelyFast} lossless compressor known for its decompresion speed; used via cBLOSC \\
        Zstd & \cite{colletFiniteStateEntropy} ZStandard lossless compressor which improves on LZ4 with better compression ratios; used via cBLOSC \\
        BLOSCLZ & \cite{altedWhyModernCPUs2010} An runtime optimized version of LZ4 in cBLOSC \\
        NONE & no compression (control case) \\
        SAMPLE NAIVE & sample every $k^{th}$ entry \\
        SAMPLE WR & sample randomly with replacement \\
        SAMPLE WOR & sample randomly without replacement \\
        FPZIP  & \cite{lindstromFastEfficientCompression2006} lossless compressor specialized for floating point \\
        SZ ABS & SZ \cite{diFastErrorBoundedLossy2016} Absolute Error Mode \\
        SZ PSNR & SZ \cite{liangErrorControlledLossyCompression2018} Peak Signal to Noise Mode \\
        SZ PW\_REL & SZ \cite{liangEfficientTransformationScheme2018} Point-Wise Relative Error Mode \\
        SZ REL & SZ \cite{diFastErrorBoundedLossy2016} Value-Range Relative Error Mode \\
        TRUNC & 64 $\rightarrow$ 32 bit float truncation \\
        ZFP ACC & ZFP \cite{lindstromFixedRateCompressedFloatingPoint2014} Fixed-Accuracy Mode \\
        ZFP PREC & ZFP \cite{lindstromFixedRateCompressedFloatingPoint2014} Fixed-Precision Mode \\
        ZFP RATE & ZFP \cite{lindstromFixedRateCompressedFloatingPoint2014} Fixed-Rate Mode \\
        \bottomrule
    \end{tabularx}
\end{table}

There are three major classes of techniques used for data reduction: dimensionality reduction which reduces the number of features, numerosity reduction which reduces the number of observations, and data compression which neither reduces the number of observations nor the number of features but instead reduces the space that the existing features and observations take in memory or storage\cite{hanDataMiningConcepts2011}.
We discuss numerosity and dimensionality reduction in Section~\ref{sec:mlai:dim_and_num} and data compression in Section~\ref{sec:mlai:background:compression}.

\subsection{Dimensionality Reduction and Numerosity Reduction}
\label{sec:mlai:dim_and_num}
The numerosity reduction class of data reduction methods reduces the number of observations in the data in either the spatial or temporal dimension.
For example, in order to control the data volume, many scientific applications or use-cases output the simulation data every $K$ time steps periodically \cite{habibHACCSimulatingSky2016, plimptonLAMMPSMolecularDynamics2004} or just store sampled data and reconstruct the full data by interpolation methods or compressed sensing technology \cite{jalaliCompressionCompressedSensing2012}. 

Since numerosity reduction is probably the most commonly used lossy compression technique used in AI/ML, it is an important benchmark for comparison.
In our study, we use 3 sampling methods: \textit{with replacement}, \textit{without replacement}, and \textit{naive sampling}.
Sampling with or without replacement is very commonly used to reduce data volumes in ML/AI applications.
The user selects a number of records at random either the same record to be retrieved multiple times (\textit{with replacement}) or only once (\textit{without replacement}) during the sampling.
\textit{Naive sampling} is a degenerate case of sampling that is still commonly used in some domains to account for large volumes of streaming data.
In the naive method, the user selects records systematically using some arbitrary method such as picking every $k^{th}$ record or the first $k$ records.
Often, naive sampling is used when the event of some limitation of the system such as having too many observations to fit in memory or too many records to process before the next arrive.

Another class of data reduction methods is dimensionality reduction techniques such as Principle Component Analysis.
These methods generally operate by projecting the data into a lower dimensional subspace with fewer features.
These techniques are orthogonal to the techniques explored in this paper because they can be used in conjunction with the later techniques we discuss.

\subsection{State-of-the-art Data Compression Techniques}\label{sec:mlai:background:compression}

Data compression is generally divided into two categories: lossless compression and lossy compression.
Lossy compression methods can now be further divided into traditional and error bounded. 
The remainder of this section is divided into three subsections: lossless methods, lossy methods, and error-bounded methods.
While the focus of our work is lossy error-bounded lossy methods, we include lossless and other lossy methods because they serve as important baselines for comparison.

\subsubsection{Lossless Compression Methods}
Lossless compression is the most widely applicable data compression method because it perfectly preserves the information through the compression and decompression process for all possible inputs.
Because of this lossless compressors such as GZIP \cite{deutschRFC1952GZIP1996} and Zstd \cite{facebookinc.ZstandardRealtimeData} have been integrated different scientific I/O libraries (such as HDF5 \cite{folkOverviewHDF5Technology2011}) 
and well-known scientific application packages (such as HACC \cite{habibHACCSimulatingSky2016} and LAMMPS \cite{plimptonLAMMPSMolecularDynamics2004}).  
Lossless compress-ability is limited by entropy\cite{shannonMathematicalTheoryCommunication1948}.
High entropy data sets like images and floating point numbers  drive researchers to look for methods that can get higher compression.

It is important to include multiple lossless compressors because they make different trade-offs between compression/decompression bandwidth and compression ratio.
We considered 3 lossless compressors that are commonly used to strike different trade-offs between bandwidth and compression ratio: 
gzip \cite{deutschRFC1952GZIP1996},
LZ4 \cite{colletLZ4ExtremelyFast},
and Zstd \cite{facebookinc.ZstandardRealtimeData}.
\textit{Gzip} is ubiquitous but is slower and less efficient than more modern approaches, with slight modifications \cite{deutschRFC1952GZIP1996}.
More modern lossless methods have been developed by Yann Collet who developed both LZ4 and Zstd.
\textit{LZ4} is known for its high decompression rate \cite{colletLZ4ExtremelyFast}.  It used a byte encoding to replace common sub-sequences of data with a compact representation similar to \cite{zivUniversalAlgorithmSequential1977}, but uses a fixed size block which improves compression time.
\textit{Zstd} builds on the concepts in LZ4 to achieve higher compression ratios at a cost of some speed.
Zstd uses LZ77 as a first stage and uses two-bit reduction phases based on finite state entropy \cite{colletFiniteStateEntropy} and Huffman encoding respectively to improve compression.

Sometimes, these methods are paired with prepossessing such as delta encoding which encodes a sequence $x_1, x_2, \dots, x_n$ as $(x_1), (x_2-x_1), (x_3-x_2), \dots, (x_n - x_{n-1})$ as used by tools like Apache Parquet.
This can increase the performance of lossless compressors by decreasing the entropy of the input sequence for data sequences whose first derivative is smoother than the raw values.

There are also specialized lossless compressors for floating point data such as  fpzip\cite{lindstromFastEfficientCompression2006}

\subsubsection{Traditional Lossy Compression Methods}
Lossy compression allows for data distortion in the reconstructed data compared with the original input and closely approximates the original data because many practical use cases do not need a ``complete facsimile'' \cite{wallaceJPEGStillPicture1992}.
Many of these methods, however, are not designed for floating point numeric data compression but rather for images with integral pixels.


One such approach commonly used in ML/AI is truncation to a lower precision.
Truncation computes a lower-order approximation of the model by neglecting the states that have a relatively low effect on the overall model. In our experiment, we have truncated the size of the data attributes from 64 IEEE bits size to 32 and 16 bits.
    
\subsubsection{Error Bounded Lossy Methods}
Error Bounded Lossy Compression (EBLC) is a recent technique that attempts to strike a balance between traditional lossless and lossy compression.
Like lossy compression, EBLC loses some information in the compression process.
However, unlike traditional measures, the user is able to specify \textit{a priori} how much error and where the error can be located in their data set with a mathematical bound on the error.
This allows users to adopt the lossy compression on their data sets with high performance and control the data distortion within the error bound.
As different programs and analyses need different concepts of error, these compressors often offer multiple kinds of error bound which preserve the data differently and need to be considered separately because they can have different effects on applications. 

There are two state-of-the-art EBLC compressors currently being used widely:  ZFP \cite{lindstromFixedRateCompressedFloatingPoint2014} and SZ \cite{diFastErrorBoundedLossy2016}.
ZFP provides 3 error bounding modes: \textit{Fixed Precision Error Bounds} (PREC), \textit{Fixed Accuracy Error Bounds} (ACC), and \textit{Fixed Rate Distortion Error Bounds} (RATE).
Fixed Precision (PREC) uses a fixed number of bit planes (number of most significant uncompressed bits) to encode each number.
Fixed Accuracy (ACC) bounds the absolute difference between a uncompressed and decompressed value.
Fixed Rate (RATE) uses a fixed number of bits to encode each value.
ZFP splits each dataset into equal-sized blocks and compresses the data block by block (the block size is set to $4^d$, where $d$ is the number of dimensions). In each block, ZFP performs an orthogonal transform on the exponent-aligned data followed by an embedded encoding algorithm. For the transformed data, ZFP truncates their insignificant bits, which is calculated based on the requested precision (i.e., different types of error bounds). Details and implementations can be found in \cite{lindstromLibraryCompressedNumerical2019}. ZFP has been widely used in many scientific applications \cite{zhangEfficientEncodingReconstruction2019}, while its performance and impact on the training in ML/AI applications are still unknown.

SZ uses various prediction methods to approximate the dataset, and falling back to lossless compression for regions of data if needed to meet a given error tolerance \cite{taoSignificantlyImprovingLossy2017, liangEfficientTransformationScheme2018, zhaoOptimizingErrorBoundedLossy2021}.
We include a subset of SZ modes: Absolute Error Bounds (ABS), Value-Range Relative Error Bounds (REL), Peak Signal to Noise Ratio (PSNR), and Point-Wise Relative Error Bounds (PW-REL).
ABS ensures that each value is with some constant of the original value.
REL  Error Bound first computes the range of values input into the compressor and keeps the bound of each below that a specified fraction of that range.
PSNR is a variant of ABS based on the PSNR metric from image compression.
Finally, PW\_REL Error Bound bounds the error as the percentage of the value.
SZ has been widely used in many scientific applications \cite{habibHACCSimulatingSky2016, cappelloUseCasesLossy2019} across different domains (cosmology, chemistry, mathematics, etc.), while its performance and impact to training data in ML/AI applications is still unknown.

\section{Problem Formalization and Methodology}
\label{sec:mlai:methodology}
The task of finding an ideal configuration for a particular application is a black-box multi-objective optimization problem.
Multi-objective because applications may be willing to trade a small amount of quality for a large increase in compression ratio or speed or visa versa.
Black-box in that one cannot observe the values of the objective functions without running an expensive computation.
Even for the small Candle dataset collecting a single observation can half an hour when running for 80 epochs.
Traditionally, solutions to multi-objective optimization problems are called Pareto points -- those that cannot improve an objective over another configuration without degrading another.
Often techniques such as evolutionary algorithms or scalarization -- using a function that expresses the user's preferences to convert the multiple objectives to a single objective -- and then use a traditional single objective approach.

However, in our case, we have knowledge about the generally expected relationships regarding parameters like error bounds and objectives such as quality and speed.
Namely, as error bounds tighten, speed tends to decrease and quality tends to increase.
We can leverage this knowledge to more quickly determine a set of candidate Pareto points.
First, observe that configurations with large error bounds that produce very low-quality results are not interesting because the user will not adopt them.
Likewise, some error-bound configurations are so small as to be indistinguishable from the original or lossless values and there exist many configurations with higher compression ratios that achieve the same quality.
Solving for these boundary conditions can be performed using existing tools such as OptZConfig \cite{underwoodOptZConfigEfficientParallel2022}.
Once we have determined these boundary conditions, we can select a number of points uniformly distributed between these boundaries based on the users' time budget.

More formally,
Table \ref{tab:notation} lists all the notations used in our paper, which can be split into two classes: compression-related parameters and analysis-related parameters. The compressor's parameters can be further split into two categories - non-fixed and fixed. The former refers to the parameters allowed to be tuned by users (such as error bound) and the latter is composed of all other parameters that are needed for running the compressor. $Q$ is a data fidelity metric function (or analysis metric) given by users for some application they are interested in (i.e. the validation accuracy for Candle-NT3). 
To discuss bandwidth, we adopt $b_n$ and $b_ c$ to represent network or storage bandwidth and compression bandwidth respectively.
The compression ratio, $\mathcal{C}$, indicates how much the volume of data has been reduced.

\begin{table}[]
    \caption{Key Notations}
    \begin{tabularx}{\columnwidth}{ lX }
    \toprule
        \textbf{Notation} & \textbf{Description} \\
    \midrule
        $d_{f,t}$ & Buffer for field, $f$, and time-step, $t$, in uncompressed form \\
        $x_i$ & $i$th element of data in $d_{f,t}$ \\
        $\vec{c}$ & Vector of nonfixed compressor parameters\\
        $\vec{\theta_c}$ & fixed compressor parameters\\
        $\tilde{d_{f,t}}(\vec{c}; \vec{\theta_{c}})$ & Decompressed buffer \\
        $\vec{\theta_m}$ & fixed parameters of the user-specified metrics function\\
        $U$ & Set of feasible nonfixed compressor parameters\\
        $\Omega$ & Set of feasible fixed and nonfixed  compressor parameters\\
        $\mathcal{Q}\left(d_{f,t},\tilde{d_{f,t}}\left(\vec{c}; \vec{\theta_c}\right); \vec{\theta_m}\right)$  &data fidelity metric function \\
        $\phi = \mathcal{Q}\left(d_{f,t}, d_{f,t}; \vec{\theta_m}\right)$ &data fidelity metric for lossless data\\
        $N$ & number of ''interesting points`` \\
        $\Lambda$ & set of ''interesting points`` (Alg.~\ref{alg:interesting})\\
        $\psi_{\vec{c}} = \mathcal{Q}\left(d_{f,t},\tilde{d_{f,t}}\left(\vec{c}; \vec{\theta_c}\right); \vec{\theta_m}\right)$  &data fidelity metric evaluated at $\vec{c}$ \\
        $\tau$ & some user-determined threshold \\
        $s_p$ & parallel speedup on $p$ cores \\
        $b_n$ & bandwidth of the network \\
        $b_c$ & bandwidth of decompression \\
        $\mathcal{C}$ & the compression ratio \\
    \bottomrule
    \end{tabularx}
    \label{tab:notation}
\end{table}

When larger values of the metric indicate better fidelity, We propose Algorithm~\ref{alg:interesting} to identify candidate Pareto points ($\Lambda$).
 There is a similar formulation where one wishes to minimize the quality metric (e.g. maximum error).
 First, we define the value of the quality metric when evaluated on a lossless configuration, as the lossless quality metric, $\phi$.
 Next, we determine using OptZConfig a configuration of the lossy compressor which has the greatest error bound which minimizes the difference from the lossless quality metric -- we call this value $\vec{u}$ the upper bound.
 After that, we determine using OptZConfig a configuration of the lossy compressor which has the greatest error bound which is still interesting to the user $\psi_{\vec{c}} < \tau$ -- we call this value $\vec{l}$ the lower bound or the minimum ``acceptable'' quality.
 
 \begin{algorithm}
 \vspace{1mm}
 \caption{Finding Candidate Pareto Points}\label{alg:interesting}
 \KwIn{$\mathcal{Q}, d_{f,t}, \tilde{d_{f,t}}, U, \vec{\theta_c},\vec{\theta_m}, \tau, N$}
 \KwOut{the set of candidate Pareto points}
 \Begin(find\_candidate\_points) {
     $\phi \gets \mathcal{Q} \left( d_{f,t}, d_{f,t} ; \vec{\theta_m} \right)$ \\
     $\vec{u} \gets \argmin_{\vec{c} \in U} \left|\phi - \psi_{\vec{c}} \right|$ \\
     $\vec{l} \gets \argmin_{\vec{c} \in U; \psi_{\vec{c}} > \tau} \psi_{\vec{c}}$ \;
     $ \Lambda  \gets \linspace\left( \vec{l}, \vec{u}, N \right)$ \tcp{numpy.linspace}
     \Return{$\Lambda$} 
 }
 \end{algorithm}

 With $\vec{l}$ and $\vec{u}$ determined, we need to sample the space in between to construct a set of interesting points.
 The number of points $N$ presents a trade-off: the larger $N$, the longer the process of measuring the points will take longer, and the smaller $N$, the less likely we are to capture the behavior between $\vec{l}$ and $\vec{u}$ faithfully.
 Users should choose the maximum $N$ allowed by their time budget.
 There are many possible ways to sample this space.
 We choose the evenly distributed points on an arithmetic scale -- what many libraries call \texttt{linspace}.
 Prior work has shown that uniform random sampling does not work well because the relationship between a compressor error bound and even simple metrics tend to have spatial properties \cite{underwoodFRaZGenericHighFidelity2020, underwoodOptZConfigEfficientParallel2022}.
 
We use these candidate Pareto points for each method in each of our experiments to examine the trade-offs between the compressors by considering the Pareto front \cite{pareto2014manual} constructed from the interesting points.

\keyfinding{Sampling the entire space exhaustively would be too expensive so instead we can use arithmetically distributed points between the $\vec{u}$ the diagnostically lossless value with the largest CR, and $\vec{l}$ the largest CR that still meets quality objectives.  We find these using a search approach\cite{underwoodOptZConfigEfficientParallel2022}}

\section{Experimental Results}
\label{sec:mlai:results}

For all our experiments, we use the hardware and software listed in Table~\ref{tab:hardware/software}.
We selected this hardware based on the availability of GPUs and CPU cores to run the models, and the availability of a high-speed network interconnect.
The software is either the system-provided default or the latest version available via the Spack package manager when the experiments were conducted.

\begin{table}
  \centering
  \caption{Hardware and Software Details}
  \begin{tabularx}{\columnwidth}{lXlX}
    \toprule
    \textbf{Component} & \textbf{Description} & \textbf{Component} & \textbf{Version}\\
    \midrule
    CPU & Intel Xeon 6148G (40 Cores) & Compiler & GCC 8.3.1 \\
    RAM & 372 GB & OS & CentOS 8 \\
    Interconnect & 100 GB/s HDR \mbox{Infiniband} & MPI & OpenMPI 4.1.0 \\
    GPU & 2 Nvidia v100 & Singularity & v3.7.1 \\
    MGARD & v0.1.0 & SZ & v2.1.12 \\
    ZFP & v0.5.5 & LibPressio & v0.72.0 \\
    fpzip & v1.3.0 & cBLOSC & v1.21.0 \\
    Python & 3.8.8 & \\
    \bottomrule
  \end{tabularx}
  \label{tab:hardware/software}
  \vspace{-.5cm}
\end{table}

In Section~\ref{sec:mlai:evaluation:breadth}, we demonstrate the broad applicability of lossy compression on several different problem classes with different kinds of solutions.
This section is concerned with the quality of the compression and the results.
Due to data and solution availability constraints, this section will use many smaller problems.
We consider variations in layout, compression principle, and the data being compressed before we scale to larger problems.
Later in Section~\ref{sec:mlai:evaluation:scalabilty}, we will show how this translates to larger problem sizes which benefit more from compression.

\subsection{Evaluating the Effect on Application Quality and Insights for Compression Development} \label{sec:mlai:evaluation:breadth}

We organize the results in this section according to our method.
First, we discuss the selection of interesting points and highlight some aspects of the compressor's performance that inform compressor selection.
Next, we broaden our focus and consider the Pareto fronts from the interesting points for several applications.
Last, we identify some insight that this can offer to compressor developers and ML/AI practitioners.

\subsubsection{Plotting Candidate Pareto Points}

\begin{figure*}[h]
    \centering
    \begin{subfigure}[t]{.30\textwidth}
        \centering
        \includegraphics[width=\linewidth]{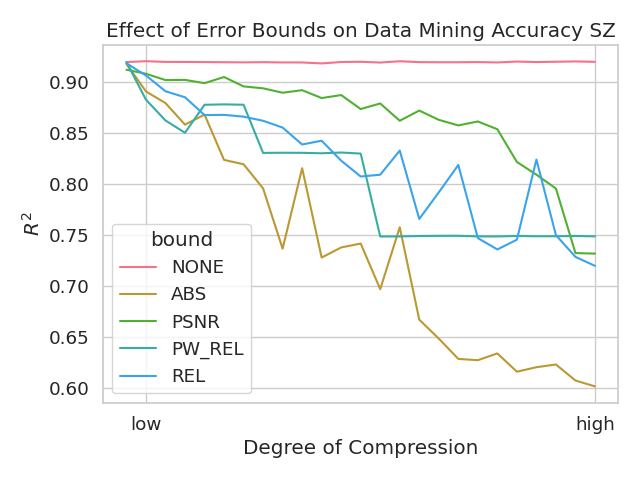}
        \caption{SZ, error bounds: ABS  = 0.1 to 1333.4, REL 1e-4 to 0.0254, PSNR 67.3 to 37.5 , PW\_REL 1e-2 to 7.9}
        \label{fig:mlai:sz_tuning}
    \end{subfigure}
    \hspace{4pt}
    \begin{subfigure}[t]{.30\textwidth}
        \centering
        \includegraphics[width=\linewidth]{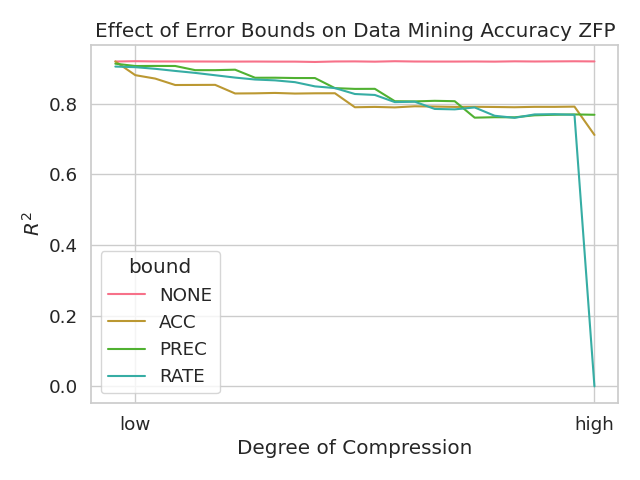}
        \caption{ZFP, error bounds: ACC 1e-5 to 2085.87, PREC 9 to 2, RATE 8.63 to 3.37}
        \label{fig:mlai:zfp_tuning}
    \end{subfigure}
    \hspace{4pt}
    \begin{subfigure}[t]{.30\textwidth}
        \centering
        \includegraphics[width=\linewidth]{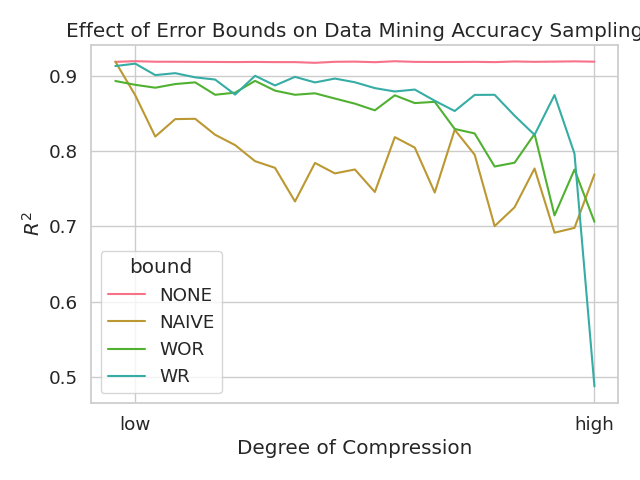}
        \caption{Sample Sizes: WOR 0.3 to 0.00625, WR 0.3 to 0.00625, Naive 1 to 73}
        \label{fig:mlai:sampling_tuning}
    \end{subfigure}%
    \caption{Tuning the Superconductor Dataset}
    \label{fig:mlai:tuning}
\end{figure*}

We begin by creating a Pareto front of the candidate Pareto points for each of the applications and methods described in this for each of the methods in Table~\ref{tab:compression_methods} and Table~\ref{tab:data-sets}.
Instead of preparing these figures for the lossless methods which do not affect quality, we simply re-ran the application several times to measure the natural variation in the quality.
Figure~\ref{fig:mlai:tuning} illustrates the results of finding candidate Pareto points for the Superconductor data sets.
While the exact ordering of the lines varies from dataset to dataset, the broad trends we describe here are consistent.

In this figure, the x-axis corresponds to increasing error bounds (i.e. lossier, generally higher compression ratios), and the y-axis represents the analysis result (i.e., Pearson's Coefficient $R^2$: the higher the better).
We selected the values that generated these bounds using \cite{underwoodOptZConfigEfficientParallel2022}.
Then, we tuned the right error bound to be the value at which the amount of error no longer increased or fell below a threshold of 70\%.

We observe candidate Pareto properties of the compressors using this algorithm.
First, as shown in Figure \ref{fig:mlai:sz_tuning}), SZ can target many levels of quality as it transitions from nearly lossless to very lossy.
This is a nice property when using these algorithms because it implies that users have a fine-grained trade-off between the level of error and compression ratio.
This is not surprising given that SZ uses a quantization-based method to bound its error such that the data distortion changes gradually with the error bound.
This kind of approach allows for fine-grain trade-offs in error as you can simply compress an increasing number of points losslessly.

Second, we note that ZFP (Figure \ref{fig:mlai:zfp_tuning}) has a threshold effect.
ZFP experiences very little loss in accuracy for a given increase in an error bound until it reaches a given level and falls rapidly.
Where this drop-off occurs is problem dependent.
This too is not particularly surprising given how ZFP operates.
ZFP achieves quality levels by selecting a number of bits to preserve.
This threshold, therefore, is likely comparable to the amount of entropy in the uncompressed data.
This property is nice in that it allows much higher levels of accuracy to be for higher levels of compression, but it does have a worrying aspect.
The value of the threshold is non-trivially related to the dataset and properties of ZFP and is difficult to predict without running the compressor multiple times.
This means that users may have a rapid loss in accuracy just by subtly increasing the error bound.

Finally, we see that Sampling (Figure~\ref{fig:mlai:sampling_tuning}) is between ZFP and SZ.
Sampling has a smoother transition than ZFP does from lossless to lossy but still features threshold effects.
It is also interesting to note that the curves are much more ``noisy" than either ZFP or SZ.
Greater levels of noise can have impacts on the performance of black-box tuning approaches to bound user metrics \cite{underwoodFRaZGenericHighFidelity2020,underwoodOptZConfigEfficientParallel2022} and deserves further study.

\keyfinding{SZ generally achieves a smoother degradation in quality as error bounds increase than sampling approaches.  ZFP had a smaller degradation until it reaches a certain (problem-dependent) degree of compression where quality collapses.  Sampling methods had a non-monotonic performance suggesting that effects can be dramatic and unexpected when applied to different datasets.  Non-monotonic behavior and threshold are behaviors to be cautious of when generalizing to other datasets for an application}

\subsubsection{Global Pareto Optimal Points}

After performing the analysis in each of the previous subsections for the remaining applications,
We consider Pareto optimal points for each of the smaller applications.
We present the results of this consideration in Figure~\ref{fig:mlai:pareto_compression}.
In these figures, the points that are best are to the upper right.

In Figure~\ref{fig:mlai:pareto_compression}, we show only the \textbf{global} Pareto points.
This means that if a compressor method does not appear in the figure, this is because it was strictly dominated by other compressors with higher quality or compression performance.
We show global Pareto fronts across all compressors to improve the figure clarity given the number of methods and datasets considered.
Some methods we tested never appeared on the Pareto front: fpzip, most sampling approaches.

The non-MLSVM results in Figure~\ref{fig:mlai:pareto_compression}, applied error bounded lossy compression independently to each column.
However, the MLSVM results  applied error-bounded lossy compression to the entire data as a matrix.
We did this in part because of differences in the design of MLSVM that made integration challenging, but also to study the impacts of different data layouts in the next subsection \footnote{Most of the analysis was conducted in Python using MPI4PY, but due to incompatibilities between MLSVM and MPI4PY required us to switch to C++ embedding the compressor into MLSVM directly where productive abstractions like NumPY multi-dimensional array broadcasting do not exist}.
We also evaluated all applications as a matrix as well, but omit them here for space because they were strictly worse for all configurations.

For some data sets, the Zstd and GZIP lossless compressor also performed well, but this was dataset dependent. Data sets with many identical values performed well with Zstd and GZIP. 
Those datasets with more varied values did poorly because of the high entropy of the mantissa bits in IEEE floating point values. 
Additionally only the highest compression level for the lossless compressors was on the Pareto front which has significant run time on most data sets, often 4 times the next closest compressor.
GZIP also required substantial time to run substantially worse than Zstd based on speed.

Sampling while a prevalent approach for data reduction in ML/AI appears on the Pareto front only once and with a substantial loss in quality in Candle NT-3.
However, given the modest improvement in compression ratio over the previous point for SZ REL and the comparatively substantial loss in quality, while Pareto optimal, this configuration may not be preferable given the comparative performance of SZ REL.

Likewise, truncation appears only once on the Pareto front for the superconductor.
While by far the fastest approach, it was frequently dominated by EBLC which achieves much larger compression ratios at similar or better quality.

Of the lossy compressors, SZ's value range relative mode and ZFP's accuracy and precision modes performed well on many data sets.
At higher allowed drops in quality, the SZ PW\_REL and PSNR modes also performed well.

\keyfinding{While different error bonding methods dominated on different applications, error-bounded lossy compressors frequently dominated the Pareto front (Figure~\ref{fig:mlai:pareto_compression}) indicating they offer the best trade-offs between application quality and compression performance oversampling and lossless compression methods widely used in ML/AI.}

\newcommand{\spacingparetofig}{.4\textwidth}
\begin{figure}
    \centering
    \begin{subfigure}{\spacingparetofig}
        \includegraphics[width=.9\textwidth]{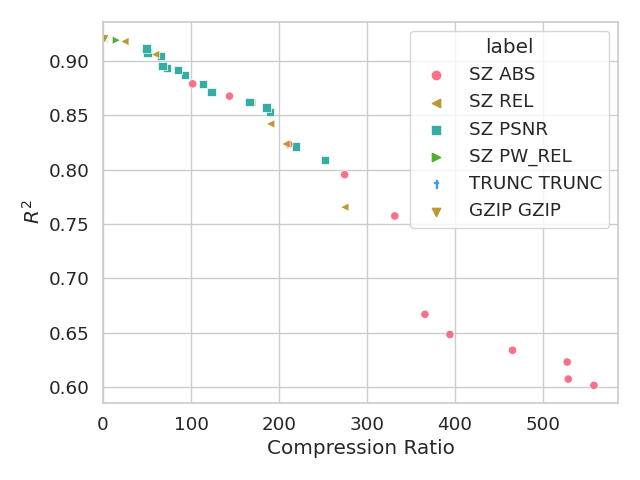}
        \caption{Superconductor}
        \label{fig:mlai:pareto_compression_superconductor}
    \end{subfigure}
    \begin{subfigure}{\spacingparetofig}
        \includegraphics[width=.9\textwidth]{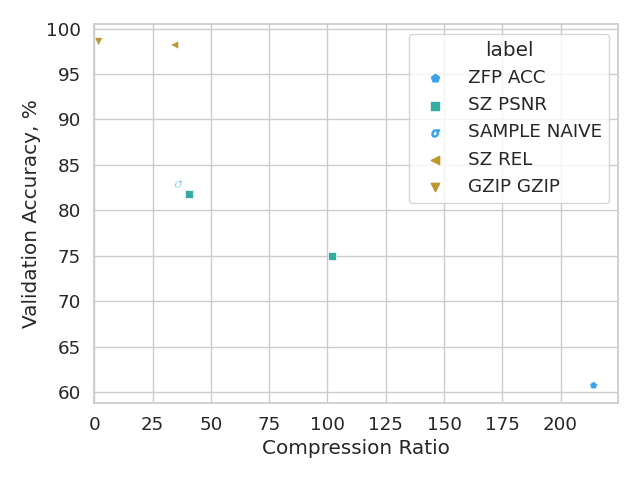}
        \caption{Candle NT-3 (small)}
        \label{fig:mlai:pareto_compression_candle}
    \end{subfigure}
    \begin{subfigure}{\spacingparetofig}
        \includegraphics[width=.9\textwidth]{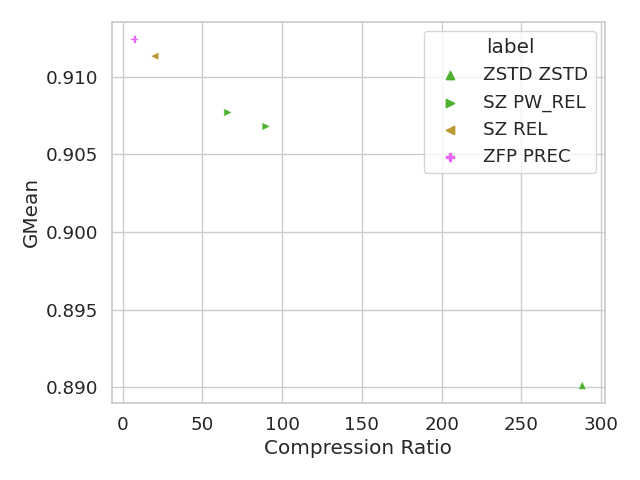}
        \caption{MLSVM Advertisement}
        \label{fig:mlai:pareto_compression_mlsvm_adver}
    \end{subfigure}
    \begin{subfigure}{\spacingparetofig}
        \includegraphics[width=.9\textwidth]{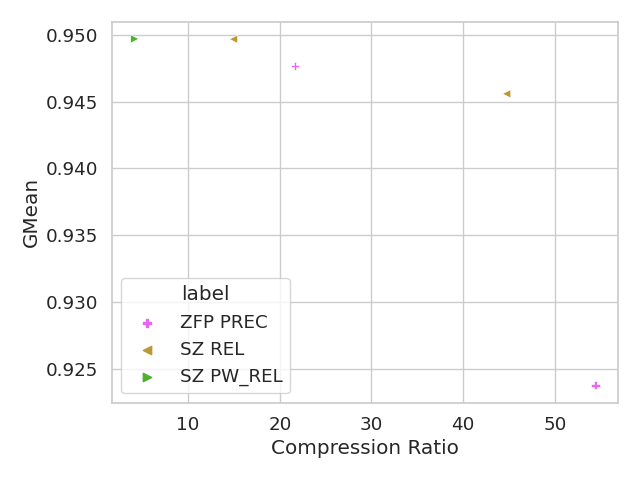}
        \caption{MLSVM Buzz}
        \label{fig:mlai:pareto_compression_mlsvm_buzz}
    \end{subfigure}
    \caption{Global Pareto Optimal Points Compression Ratio and Quality for Various Applications.  Methods omitted when not pareto optimal}
    \label{fig:mlai:pareto_compression}
\end{figure}

\subsubsection{What insights does this provide for ML/AI practitioners and compressor designers}

We found that independent column-based lossy compression is most effective at preserving quality for ML/AI problems with inputs structured like these.
This is because the values in each column are more likely to be spatially correlated with each other where as the contributions of rows are reduced by the accuracy of transforms and predictions.
This has slight size overhead for compressors such as SZ where redundant metadata (i.e. Huffman trees) are stored for each column regardless of input size and run-time performance overhead from repeated memory allocation latency when columns are small.
In the future, a specialized version could alleviate the overheads.

We also tried applying prepossessing techniques such as delta encoding and double delta encoding by column to the input data to attempt to improve the lossless compression performance (cf. Apache Parquet). However, we did not see an improvement for these data sets; we instead observed a drop in compression ratio from $\approx 2 \times$ to about $1.5 \times$  for all configurations and compressors.  This can be accounted for by the non-smoothness of the adjacent values in the columns.

\begin{figure}
    \centering
    \includegraphics[width=\columnwidth]{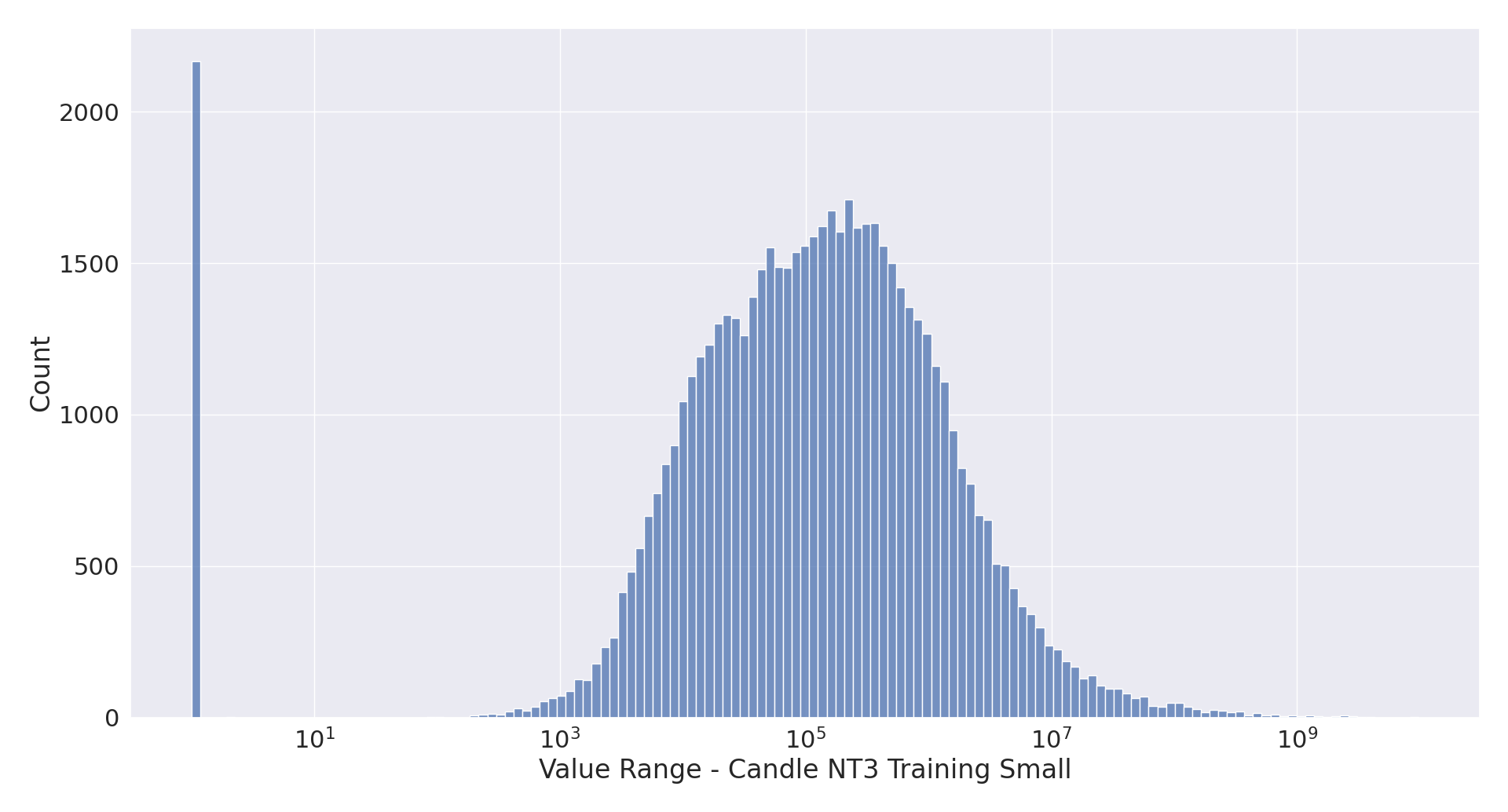}
    \caption{Distribution of the value Ranges for Candle NT-3.  A wide array of values can be observed}
    \label{fig:mlai:histranges}
\end{figure}

We have found that prediction-based error-bounded lossy compression was generally the most effective.
Even though the compressor modes in SZ share substantial amounts of code, the value range relative mode was most effective when compressing by column. 
All modes were similarly effective when compressing as a matrix.
This is because using value range relative mode by column effectively set distinct error bounds for each column.
Figure~\ref{fig:mlai:histranges} shows the distribution of the value ranges by column for Candle NT-3\footnote{There is a spike at value ranges of 0 indicating no-variance.  We used the application as provided and did not correct for features like this.}.
We expect that ZFP would have similar benefits if an adapter was used for ZFP to preserve a value-range relative bound.
Sampling methods had erratic quality effects (e.g. sharp drops, and non-monotonic behavior) and were also not often on the Pareto front being dominated by either quality or compression performance.

\keyfinding{Value range relative error bounds applied by column generally results in the best compression for tabular datasets and quality performance trade-offs because datasets value ranges can vary greatly by feature even improving against sampling and other techniques}

\subsection{Performance Evaluation} \label{sec:mlai:evaluation:scalabilty}
To evaluate the performance implications of using lossy compression, there are, in principle, two possible slowdowns.
One possibility is that training time would change, we found this was not true for the applications we tested.
The other possibility is the impact on IO time, this is the focus of the remainder of this section.

To evaluate the scalability of using error-bounded lossy compression, we will use larger scientific data sets.
We have two such larger data sets: Candle-NT3 and PtychoDNN.
For each application, we generated a 4GB binary input file \footnote{Candle NT-3's native input format is CSV, however, to mitigate the large overheads of parsing the CSV into a binary form for the compressors and processing by Candle. All input sizes reported for Candle assume this binary format instead of the native CSV format; Our results are even stronger when compared against CSV with parsing due to this overhead. PtychoDNN uses numpy format which is a binary format with a small header and thus did not require conversion} using the tools provided by the applications.
Note that this generalizes to even larger data sets that are processed in chunks; this is to overcome maximum input sizes for some compressors such as LZ4 and Zstd \cite{colletLZ4ExtremelyFast,facebookinc.ZstandardRealtimeData}.

The run-time without compression as the time to read the entire data over the WAN:  $ \frac{||d_{f,t}||}{b_n} $ One can model the time used to perform read the data with compression can be modeled as the sum of the transfer time in compressed format + the parallel decompression time:  $ \frac{||d_{f,t}||}{s_p b_c} + \frac{||d_{f,t}||}{\mathcal{C} b_n} $.  Thus the speedup is:
 $ \frac{\mathcal{C} s_p b_c}{\mathcal{C} b_n + s_p b_c}$.  We can then compute the number of cores required to achieve a parallel speedup:  $ s_p > \frac{\mathcal{C} b_n}{b_c \left( \mathcal{C} -1 \right)} $.
WAN bandwidth can vary widely from site to site, and across different times of day.
Therefore, we compute the number of parallel compressions required to achieve a speedup with varying IO latency.

We consider 3 speeds for the literature: 
 3.75 GB/s  -- the speed of the dedicated link between the Advanced Photon Source and Argonne Leadership Computing Facility at Argonne National Laboratory \cite{salimBalsamRealTimeExperimental2019};
1 GB/s -- the typical speed that users can transfer data using Globus \cite{kettimuthuTransferringPetabyteDay2018} \footnote{Greater speeds are possible as described in this paper, but requires cooperation with the Internet Service Providers and other optimizations not available to typical users};
and 125 MB/s -- the typical speed for broadband internet at remote locations such as for intelligent transportation systems in rural communities\cite{rahmanDynamicErrorboundedLossy}.
Now we must determine the bandwidth of the compressors.
The bandwidth of the compressors depends on the dataset and the configuration of the compressors.
In Tables~\ref{tab:coresforspeedup_candle} and~\ref{tab:coresforspeedup_ptychonn}, we show the bandwidth achieved using the SZ compressor in value-range-relative mode -- one of the compressors which performed the best in Section~\ref{sec:mlai:evaluation:breadth} for the Candle-NT3 and Ptychonn data-sets.
We further show multiple error bounds to represent differing levels of quality requirements depending on the user.

For even high-performing networks, speedups can be accomplished by transporting the compressed data and then decompressing the data using CPU parallel versions of the compressors using a modest number of cores.
GPU parallel versions of compressors currently require additional improvements to compete with parallel CPU compressors due in large part to the time to transfer the data to the GPU \cite{underwoodOptZConfigEfficientParallel2022} but that may change with further improvements to these compressors.

\keyfinding{For each network bandwidth, we can find a reasonable degree of parallelism that accelerates compression for training data transfers}

\begin{table}[]
    \centering
    \caption{Parallel Speedup Threshold: SZ-REL/Candle NT-3}
    \label{tab:coresforspeedup_candle}
    \begin{tabularx}{\columnwidth}{llllrrr}
\toprule
   psnr   &        bound ($\vec{c}$) & $\mathcal{C}$ &  $b_c$ & cores & for a &  speedup \\
   &    &  &   & 3.75 GB/s & 1GB/s & 125 MB/s \\
   \midrule
164.7 &1e-8       &    5.4  &              0.17        &  27  & 8 &  1   \\ 
144.7 &1e-7       &    7.6  &              0.28        &  16  & 5 &  1\\ 
124.7 &1e-6       &   13.1  &              0.40        &  11  & 3 &  1\\ 
104.1 &1e-5       &   35.2  &              0.71        &   6  & 2 &  1\\ 
 93.0 &1e-4       &  201.5  &              1.28        &   3  & 1 &  1\\ 
 83.8 &1e-3       & 1476.6  &              1.44        &   3  & 1 &  1\\
  \bottomrule
    \end{tabularx}
\end{table}

\begin{table}[]
    \centering
    \caption{Parallel Speedup Threshold SZ-REL/Ptychonn}
    \label{tab:coresforspeedup_ptychonn}
\begin{tabularx}{\columnwidth}{rrrrrrr}
\toprule
   psnr   &        bound ($\vec{c}$) & $\mathcal{C}$ &  $b_c$ & cores & for a &  speedup \\
   &    &  &   & 3.75 GB/s & 1GB/s & 125 MB/s \\
\midrule
164.90 & 1e-08 &               2.22 &                     0.06 &             122 &             33 &                5 \\
144.80 & 1e-07 &               3.05 &                     0.08 &              69 &             19 &                3 \\
124.87 & 1e-06 &               4.61 &                     0.13 &              38 &             11 &                2 \\
105.93 & 1e-05 &              10.67 &                     0.21 &              21 &              6 &                1 \\
 90.71 & 1e-04 &              40.46 &                     0.38 &              11 &              3 &                1 \\
\bottomrule
\end{tabularx}
\end{table}

\section{Related Work} \label{sec:mlai:related_work}

In this section, we discuss the related work in two facets: lossy compression in ML/AI training and the lossy compression design used elsewhere in ML/AI.

\subsection{Lossy Compression of ML/AI training data}
There have been several attempts to apply lossy compression to data mining in the past.
These works have almost exclusively focused on images.
As explained in the introduction, images differ in their structure than scientific tabular data (e.g. limited to no spatial correlation) and may not be comparable.
%
%
The paper by \cite{jalaliCompressionCompressedSensing2012} considers the related domain of compressive sensing.
They argue that for some lossy compressors such as JPEG with a given rate distortion, it is possible to approximately lower bound the number of observations required to "successfully" reconstruct the original signal.
Thus approximately "bounding" certain classes of lossy compressors are commonly used in image storage.

Some papers use very early forms \cite{jagadishSemanticCompressionPattern1999} of lossy compression in pattern mining \cite{garofalakisDataMiningMeets}.
Essentially, the authors construct a decision tree for a subset of columns called predicted columns. If these trees accurately predict within the error bound and the decision tree is smaller, it is stored instead.
However, these techniques were expensive to execute and have a chance of not compressing at all because of the error bound.
They also don't attempt to quantify the effect of the loss on the decisions made on the decompressed data.
%
%

Joseph et al\cite{josephCorrectnesspreservingCompressionDatasets2020} is the most similar, but we make several notable improvements.
Their work focuses on images; we focus on scientific tabular data.
Image data differs in that it has spatial features whereas scientific data may be uncorrelated.
The only find that ZFP is safe for use in training data images in ML/AI.
We go much further in identifying the value of value range relative error bounds for tabular data, demonstrating the EBLC compressor's superiority over widely adopted sampling approaches, and show that other EBLCs perform better than ZFP and explain why.
In total, we use 17+ data reduction schemes including error-bounded lossy (EBLC) whereas they use only 3.
The value of including many additional methods is that we can 1) show the comparative benefits of different approaches relative to current practice 2) we were able to include new methods including  SZ which actually vastly outperforms ZFP on these applications which was the best-performing method in \cite{josephCorrectnesspreservingCompressionDatasets2020}.
Additionally, We also show that using a value range relative error bound by column works well for tabular datasets -- an error bound not natively supported by any existing compressor --and propose a multi-objective search method for black box compressors that work with all compressors \cite{underwoodOptZConfigEfficientParallel2022} to automatically identify the tradeoffs between quality and speed or quality and compression ratio.
These contributions will both improve how applications are evaluated using lossy compressed inputs and improve the future of compressor designs.

\subsection{Lossy Compression for ML/AI more broadly}
There have been a few papers that have looked at particular aspects of applying lossy compression to ML/AI applications.

Another example, the paper \cite{evansJPEGACTAcceleratingDeep2020} considers the effect of compressing activation data for deep neural networks using JPEG image compression as well as  Precision Reduction, Run Length encoding, and Zero Value Compression compression.
Like this paper, ours considers the application of lossy compression, but the focus is different: they focused on activation data for deep neural networks whereas we focus on the compression of training and validation data for a greater variety of methods to provide a method for evaluating the impacts of lossy compression on ML/AI applications.
Additionally, the paper by Evans et al considers only two lossy compressors and neither of them is particularly modern or designed for scientific applications.

\cite{jinNovelMemoryEfficientDeep2020} considers the impacts of  a GPU version of SZ version on the gradients used in convolutional DNNs.
Unlike  \cite{evansJPEGACTAcceleratingDeep2020} which was primarily empirical, the paper by Jin et al uses an analytical approach that makes several simplifying assumptions that are true for the applications they consider and cuSZ compressor specifically.
Similar work has been performed by~\cite{idelbayevLCFlexibleExtensible2021} but for a bespoke compressor.
One of these assumptions is that the error distribution imposed by the compressor is uniform; however, this is not true of many of the leading lossy compressors including newer versions of SZ \cite{lindstromErrorDistributionsLossy2017}.
Further, their formulation is also limited in that applies only to networks whose gradient is just the sum of the product of the activation data and loss -- i.e. not containing any dropout or other more complex operators.
Whereas this work considers many more variations.

\section{Conclusions and Future Work}\label{sec:mlai:conclusions}

We present a systematic and automated methodology for evaluating  data  reduction techniques  for  ML/AI.  
We then  use  it  to  perform  a  much  more  comprehensive  evaluation than has previously been attempted.
We consider  17  data reduction methods on 7 ML/AI applications to show that modern lossy compression methods can achieve a 50-100$\times$ improvement in compression ratio for a 1\% or less loss in quality.
By considering such a wide array of applications and error-bounded compression methods we show the broad applicability of these methods to a variety of applications relative to the various baselines used in the community demonstrating that for many applications error-bounded lossy compression can optimize workflows with minimal impact on application quality.

We also identified several novel, key insights not offered by previous studies that will enable future studies and guide the adoption of lossy compression in ML/AI:
1) we proposed an efficient method to search for Pareto points in the configuration of lossy compression where quality is a parameter.
This method allows users to quickly and automatically get an initial evaluation of the trade-offs of applying lossy compression on their application without the overhead of exhaustive evaluation.
2), we show by plotting the candidate Pareto points some kinds of generalization pitfalls that can be encountered during lossy compression -- namely threshold and non-monotonic behavior -- that can indicate
3), we identified that compressing with value relative error bounds by column in tabular datasets results in a higher quality for these applications because value ranges can vary greatly by feature.
No compressor currently implements these bounds natively and implementing them on top of existing compressors drives up space and time overhead unnecessarily providing an opportunity to improve lossy compressor design.
4), we demonstrate that even on high-performance networks, there are opportunities to improve performance by using lossy compression to accelerate I/O.


%
%
\ifCLASSOPTIONcaptionsoff
  \newpage
\fi

\bibliography{sample-base}
\bibliographystyle{IEEEtran}

\end{document}